\documentclass[10pt,twocolumn,letterpaper]{article}

\usepackage{cvpr}
\usepackage{times}
\usepackage{epsfig}
\usepackage{graphicx}
\usepackage{amsmath}
\usepackage{amssymb}
\usepackage{courier}
\usepackage{amsmath}
\usepackage{amssymb}
\usepackage{amsfonts}
\usepackage{indentfirst}
\usepackage{xcolor}
\usepackage{multirow}
\usepackage{float}
\usepackage{url}
\usepackage{graphicx}
\usepackage{graphics}
\usepackage{subfigure}
\usepackage{float}
\usepackage{bigstrut}
\usepackage{booktabs}
\usepackage{bibentry}
\usepackage{epstopdf}
\usepackage{tabularx}
\usepackage{balance}
\usepackage{verbatim}

 \makeatletter
    \newcommand{\thickhline}{%
        \noalign {\ifnum 0=`}\fi \hrule height 1pt
        \futurelet \reserved@a \@xhline
    }
    \newcolumntype{"}{@{\vrule width 1pt}}
    \makeatother

\cvprfinalcopy 

\def\cvprPaperID{158} 

\ifcvprfinal\fi
\begin{document}

\title{Modeling Temporal Dynamics and Spatial Configurations of Actions Using Two-Stream Recurrent Neural Networks}

\pdfinfo{
/Title (Modeling Temporal Dynamics and Spatial Configurations of Actions Using Two-Stream Recurrent Neural Networks)
/Author (Hongsong Wang, Liang Wang)
/Subject (2017 IEEE Conference on Computer Vision and Pattern Recognition)
}

\author{Hongsong Wang$^{1,3}$ \quad  \quad  Liang Wang$^{1,2,3}$\\
$^1$Center for Research on Intelligent Perception and Computing (CRIPAC),\\
National Laboratory of Pattern Recognition (NLPR)\\
$^2$Center for Excellence in Brain Science and Intelligence Technology (CEBSIT),\\
Institute of Automation, Chinese Academy of Sciences (CASIA)\\
$^3$University of Chinese Academy of Sciences (UCAS)\\
{\tt\small \{hongsong.wang, wangliang\}@nlpr.ia.ac.cn}
}

\maketitle
\thispagestyle{empty}

\begin{abstract}
Recently, skeleton based action recognition gains more popularity due to cost-effective depth sensors coupled with real-time skeleton estimation algorithms. Traditional approaches based on handcrafted features are limited to represent the complexity of motion patterns. Recent methods that use Recurrent Neural Networks (RNN) to handle raw skeletons only focus on the contextual dependency in the temporal domain and neglect the spatial configurations of articulated skeletons. In this paper, we propose a novel two-stream RNN architecture to model both temporal dynamics and spatial configurations for skeleton based action recognition. We explore two different structures for the temporal stream: stacked RNN and hierarchical RNN. Hierarchical RNN is designed according to human body kinematics. We also propose two effective methods to model the spatial structure by converting the spatial graph into a sequence of joints. To improve generalization of our model, we further exploit 3D transformation based data augmentation techniques including rotation and scaling transformation to transform the 3D coordinates of skeletons during training. Experiments on 3D action recognition benchmark datasets show that our method brings a considerable improvement for a variety of actions, i.e., generic actions, interaction activities and gestures.
\end{abstract}

\section{Introduction} \label{introduction}
Human action recognition \cite{aggarwal2011human} has become an active area in computer vision and there are many important research problems, such as event recognition \cite{jiang2013high}, group based activities recognition \cite{lan2010beyond}, human object interactions \cite{gupta2009observing} and activities in egocentric videos \cite{li2015delving,fathi2012learning}.
Most approaches have been proposed to recognize actions in RGB videos recorded by 2D cameras. However, it still remains a challenging problem for three reasons. First, it is hard to well extract useful information from the high dimensional and low quality input data. Second, the RGB video is highly sensitive to some factors like illumination changes, occlusion and background clutter. Third, the identification of actions is related to high-level visual clues such as human poses and objects, which are very difficult to obtain from RGB videos directly.

\begin{figure}
\centering
\includegraphics[width=1.0\linewidth]{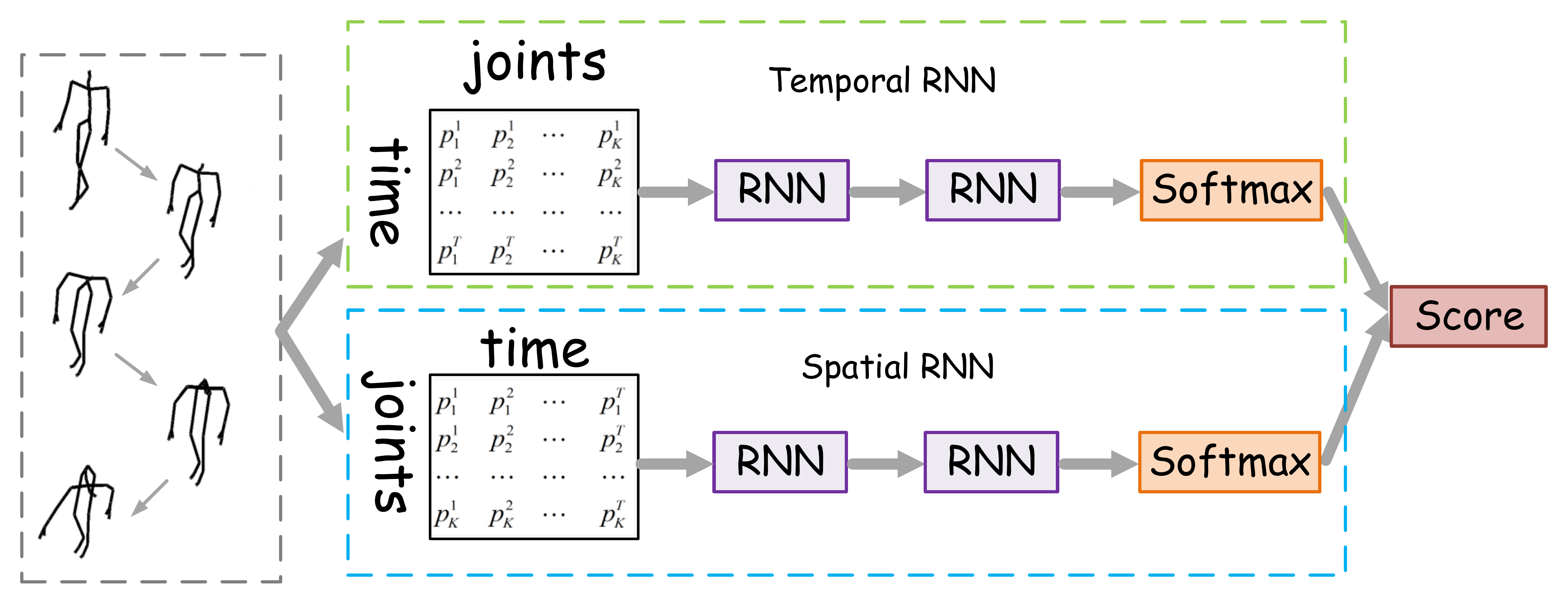}
\caption{A two-stream RNN architecture for skeleton based action recognition. Here \emph{Softmax} denotes a fully connected layer with a softmax activation function.} 
\label{fig:two_stream}
\vspace{-2mm}
\end{figure}

Humans can recognize actions with a few spots describing motions of the main joints of skeletons \cite{johansson1973visual}, and experiments show that a large set of actions can be recognized solely from skeletons \cite{johansson1975visual}. 
In contrast to RGB based action recognition, skeleton based action recognition can avoid the awful task of feature extraction from videos and explicitly model the dynamics of actions. There are three ways to obtain skeletons: motion capture systems, RGB images and depth maps. Sophisticated motion capture systems are very expensive and require the user to wear a motion capture suit with markers. Extracting reliable skeletons from monocular RGB images or videos, i.e., pose estimation, is still an unsolved problem. Fortunately, with the recent advent of affordable depth sensors, it is much easier and cheaper to obtain 3D skeletons from depth maps. For example, Shotton et al. \cite{shotton2013real} propose a method to quickly and accurately predict 3D positions of body joints from a single depth image. These advances excite considerable interest for skeleton based action recognition and various algorithms have been proposed recently.

Traditional skeleton based action recognition approaches are mainly divided into two categories: joint based approaches and body part based approaches. Joint based approaches consider the human skeleton as a set of points and use various positions based features such as joint positions \cite{hussein2013human,lv2006recognition} and pairwise relative joint positions \cite{wang2012mining,yang2012eigenjoints} to characterize actions. While body part based approaches regard the human skeleton as a connected set of segments, and then focus on individual or connected pairs of body parts \cite{yacoob1998parameterized} and joint angles \cite{ohn2013joint}. Based on handcrafted low-level features, both approaches employ relatively simple time series models, e.g., hidden Markov model \cite{wu2014leveraging,xia2012view}, to recognize actions. However, human-engineered features are limited to represent the complexity of the intrinsic characteristics of actions and the subsequent time series models do not unleash the full potential of the sequential data.

Inspired by the great success of deep learning for RGB based action recognition \cite{simonyan2014two,karpathy2014large,ji20133d}, there is a growing trend of using deep neural networks for skeleton based action recognition.
Different structures of Recurrent Neural Networks (RNN), e.g., hierarchical RNN \cite{du2015hierarchical}, RNN with regularizations \cite{zhu2016co}, differential RNN \cite{veeriah2015differential} and part-aware Long Short-Term Memory (LSTM) \cite{shahroudy2016ntu}, have been used to learn motion representations from raw skeletons.
However, considering an action is a continuous evolution of articulated rigid segments connected by joints \cite{zatsiorski1998kinematics}, these RNN-based methods only model the contextual information in the temporal domain by concatenating skeletons for each frame. In fact, different actions are performed with different spatial configurations of joints of skeletons. The dependency in the spatial domain also reflects the characteristics of actions and should not be neglected for skeleton based action recognition.

To this end, we introduce a novel two-stream RNN architecture which incorporates both spatial and temporal networks for skeleton based action recognition. Figure \ref{fig:two_stream} shows the pipeline of our method. The temporal stream uses a RNN based model to learn the temporal dynamics from the coordinates of joints at different time steps. We employ two different RNN models, \emph{stacked RNN} and \emph{hierarchical RNN}. Compared with \emph{stacked RNN}, \emph{hierarchical RNN} is designed according to human body kinematics and has fewer parameters. At the same time, the spatial stream learns the spatial dependency of joints. We propose a simple and effective method to model the spatial structure that first casts the spatial graph of articulated skeletons into a sequence of joints, then feeds this resulting sequence into a RNN structure. Different methods are explored to turn the graph structure into a sequence for the purpose of better maintaining the spatial relationships.
The two channels are then combined by late fusion and the whole network is end-to-end trainable. Finally, to avoid overfitting and improve generalization, we exploit data augmentation techniques by using 3D transformation, i.e., rotation transformation, scaling transformation and shear transformation to transform the 3D coordinates of skeletons during training.

In summary, the main contributions of this paper are listed as follows.
First, we propose a two-stream RNN architecture to utilize both spatial and temporal relations of joints of skeletons. Second, we exploit and compare different architectures of both streams. Third, we propose data augmentation techniques based on 3D transformation and demonstrate the effectiveness for skeleton based action recognition.
Finally, our method obtains the state-of-the-art results on three important benchmarks for a variety of actions, i.e., generic actions (NTU RGB+D), interaction activities (SBU Interaction) and gestures (ChaLearn).

\section{Related work}
In this section, we briefly review action recognition approaches related to ours. The two aspects are as follows.

\subsection{Action recognition with deep networks}
Deep neural networks have made great progress in the area of action recognition. 3D Convolutional Neural Networks (CNN) is proposed and different architectures are studied to take advantage of local spatio-temporal information \cite{karpathy2014large,ji20133d}. To capture complementary information between appearance and motion, a two-stream CNN architecture is developed for RGB based action recognition \cite{simonyan2014two}.

Recently, Recurrent Neural Networks (RNN) have been widely used for action recognition. Srivastava et al. \cite{srivastava2015unsupervised} use multilayer Long Short Term Memory (LSTM) networks to learn representations of video sequences. Donahue et al. \cite{donahue2014long} develop an end-to-end trainable Long-term Recurrent Convolutional Networks (LRCN) architecture which can simultaneously learn temporal dynamics and convolutional perceptual representations from RGB videos. Deep Convolutional and Recurrent Neural Networks has also been proposed and applied for activity recognition \cite{ordonez2016deep,huang2015bidirectional}.

Prior to our work, several models have been proposed based on RNN for skeleton based action recognition. Du et al. \cite{du2015hierarchical,du2016representation} first design an end-to-end hierarchical RNN architecture for skeleton based action recognition. Zhu et al. \cite{zhu2016co} propose a fully connected deep LSTM network with regularization terms to learn co-occurrence features of joints. Veeriah et al. \cite{veeriah2015differential} present differential RNN that extends LSTM structure by modeling the dynamics of states evolving over time. Shahroudy et al. \cite{shahroudy2016ntu} propose a part-aware extension of LSTM to utilize the physical structure of the human body. These methods only model the motion dynamics in the temporal doamin and neglect the spatial configurations of articulated skeletons. Recently, Liu et al. \cite{liu2016spatio} extend LSTM to spatial-temporal domain for the purpose of modeling the dependencies between joints. As temporal dynamics and spatial configurations are separate visual pathways \cite{goodale1992separate}, we employ a two-stream architecture to model them accordingly. 

\subsection{Features based on skeletons}
Previous skeleton based action recognition methods mainly focus on handcrafted features \cite{aggarwal2014human}. To get representations of postures, one straightforward feature is the pairwise joint location difference, which can be simply concatenated \cite{masood2011measuring}, or casted into 3D cone bins to build a histogram of 3D joints locations \cite{xia2012view} for action recognition.

Joint orientation is another good feature as it is invariant to the human body size. For example, Sempena et al. \cite{sempena2011human} apply dynamic time warping based on the feature vector built from joint orientation along time series. Bloom et al. \cite{bloom2012g3d} use AdaBoost to combine five types of features, i.e., pairwise joint position difference, joint velocity, velocity magnitude, joint angle velocity and 3D joint angle to recognize gaming actions, for real-time action recognition.

There are some work that groups the joints of skeletons to construct planes from joints and then measures the joint-to-plane distance and motion. Yun et al. \cite{yun2012two}  capture the geometric relationship between the joint and the plane spanned by three joints. Sung et al. \cite{sung2011human} compute the joint's rotation matrix w.r.t. the person's torso, hand position w.r.t. the torso and joint rotation motion as features.

\section{Overview of RNN}
Different from feedforward neural networks that map from one input vector/matrix to one output vector/matrix, recurrent neural networks (RNN) map an input sequence \(X\) to another output sequence \(Y\).

\begin{figure}
\centering
\includegraphics[width=0.9\linewidth]{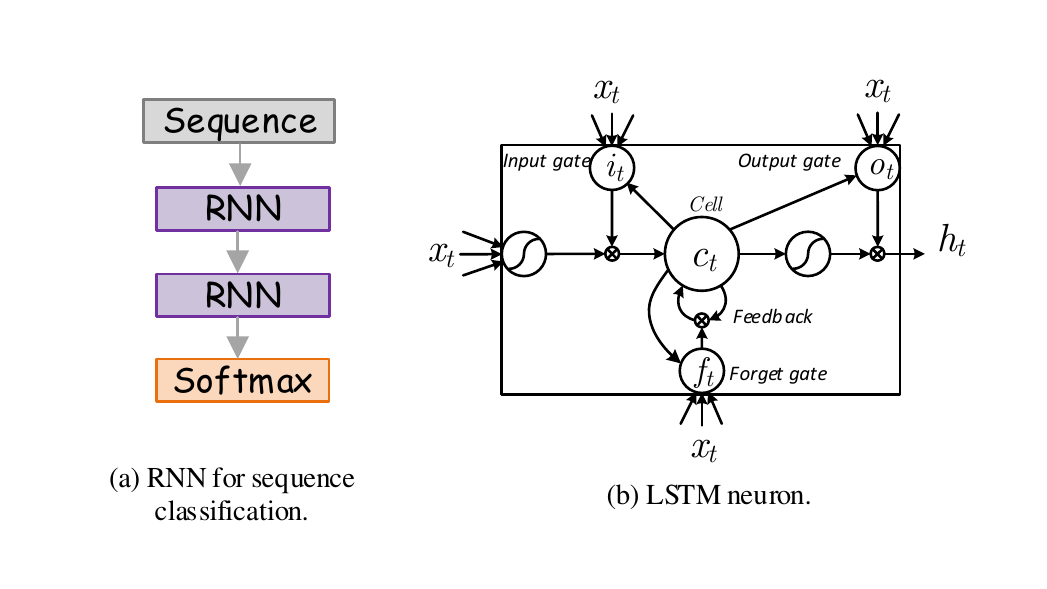}
\caption{(a) A two-layer stacked RNN for sequence classification. (b) A LSTM block with input, output, and forget gates \cite{hochreiter1997long}.}
\label{fig:lstm_unit}
\vspace{-2mm}
\end{figure}

RNN architectures are naturally suitable for the sequence classification, where each input sequence is assigned with a single class. Layers of RNN can be stacked to build a deep RNN by considering the output sequence of the previous layer as the input sequence of the current layer. A typical structure of RNN for sequence classification is shown in Figure \ref{fig:lstm_unit}(a), which contains a stack of RNN layers with a softmax classification layer on top of the last hidden layer.

Due to the vanishing gradient and error blowing up problems \cite{hochreiter2001gradient}, the standard RNN cannot store information for long periods of time or access the long range of context. Long short-term memory (LSTM) \cite{hochreiter1997long} addresses this problem by using additional gates to determine when the input is significant enough to remember, when it should continue to remember or forget the value, and when it should output the value. The LSTM unit has been shown to be capable of storing and accessing information over very long timespans \cite{gers2002learning}. Figure \ref{fig:lstm_unit}(b) depicts a LSTM unit:
\begin{equation}
\begin{array}{l}
{i_t} = \sigma ({W_{xi}}{x_t} + {W_{hi}}{h_{t - 1}} + {W_{ci}}{c_{t - 1}} + {b_i})\\
{f_t} = \sigma ({W_{xf}}{x_t} + {W_{hf}}{h_{t - 1}} + {W_{cf}}{c_{t - 1}} + {b_f})\\
{c_t} = {f_t}{c_{t - 1}} + {i_t}\tanh ({W_{xc}}{x_t} + {W_{hc}}{h_{t - 1}} + {b_c})\\
{o_t} = \sigma ({W_{xo}}{x_t} + {W_{ho}}{h_{t - 1}} + {W_{co}}{c_t} + {b_o})\\
{h_t} = {o_t}\tanh ({c_t})
\end{array}
\end{equation}
where \(i,f,o\) correspond to the input gate, forget gate and output gate, respectively. All the matrices \(W\) are the connection weights and all the variables \(b\) are biases.

\section{Two-stream RNN}
The sequence of skeletons determines the evolution of actions, which has both spatial and temporal structures. The spatial structure displays a spot of the pictorial form of joints while the temporal structure tracks and represents the movement of joints. Accordingly, we devise an end-to-end two-stream architecture based on RNN, which is shown in Figure \ref{fig:two_stream}. Here the fusion is performed by combining the softmax class posteriors from the two nets.

\subsection{Temporal RNN}
We begin with the description of the temporal channel of RNN, which models the temporal dynamics of skeletons. Similar to the previous work \cite{du2015hierarchical,zhu2016co,veeriah2015differential,shahroudy2016ntu}, it concatenates the 3D coordinates of different joints at each time step and handles the generated sequence with a RNN architecture. We focus on the following two model structures.

\noindent \textbf{Stacked RNN.} This structure feeds the RNN network with the concatenated coordinates of all joints at each time step. Here we stack two layers of RNN and find that adding more layers would not considerably improve the performance. As the length of skeleton sequences is relatively long (e.g., 50$\scriptsize{\sim}$200), we adopt LSTM neurons for all layers. Although simple, \emph{stacked RNN} has been widely used to process and recognize sequences of variable lengths.

\begin{figure}
\centering
\includegraphics[width=0.9\linewidth]{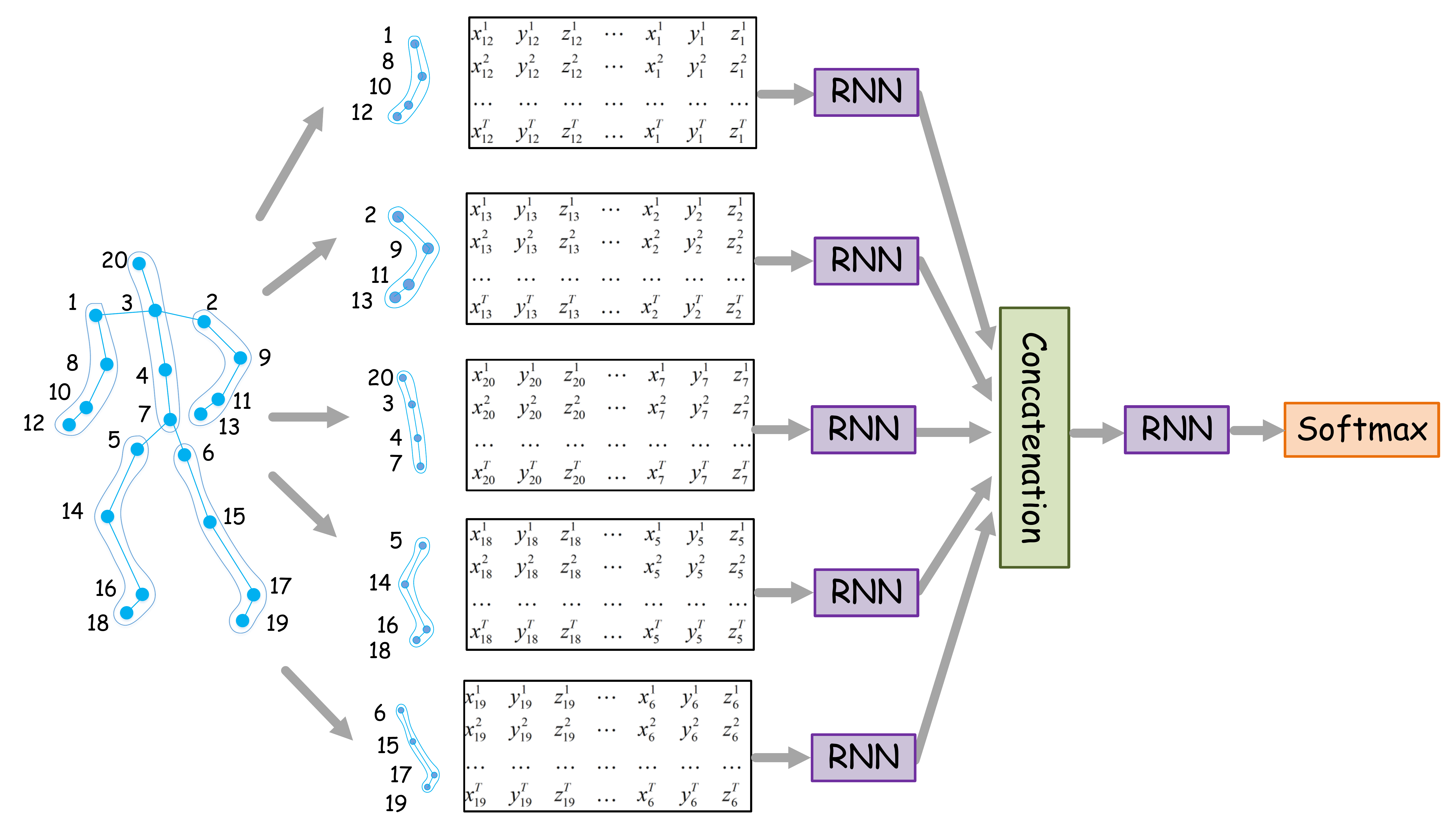}
\caption{Hierarchical RNN for skeleton based action recognition.}
\label{fig:hier_rnn}
\vspace{-2mm}
\end{figure}

\noindent \textbf{Hierarchical RNN.} The human skeleton can be divided into five parts, i.e., two arms, two legs and one trunk. We observe that an action is performed by either an independent part or a combination of several parts. For example, kicking depends on legs and running involves both legs and arms. Thus, a hierarchical structure of RNN is used to model the motions of different parts as well as the whole body. Figure \ref{fig:hier_rnn} shows the proposed structure. To be consistent with the \emph{stacked RNN} structure, our \emph{hierarchical RNN} also has two layers vertically.

In the first layer, we use a corresponding RNN to model the temporal movement of each body part based on its concatenated coordinates of joints at each time step.
In the second layer, we concatenate the outputs of the RNN of different parts and adopt another RNN to model the movement of the whole body. Compared with the pioneered hierarchical structure in \cite{du2015hierarchical}, our structure is more succinct and straightforward, and does not use additional fully connected layers before the logistic regression classifier with softmax activation. Compared with the stacked structure, the hierarchical structure has relatively fewer parameters and is less likely to overfit.

\subsection{Spatial RNN}
\begin{figure}
\centering
\includegraphics[width=0.9\linewidth]{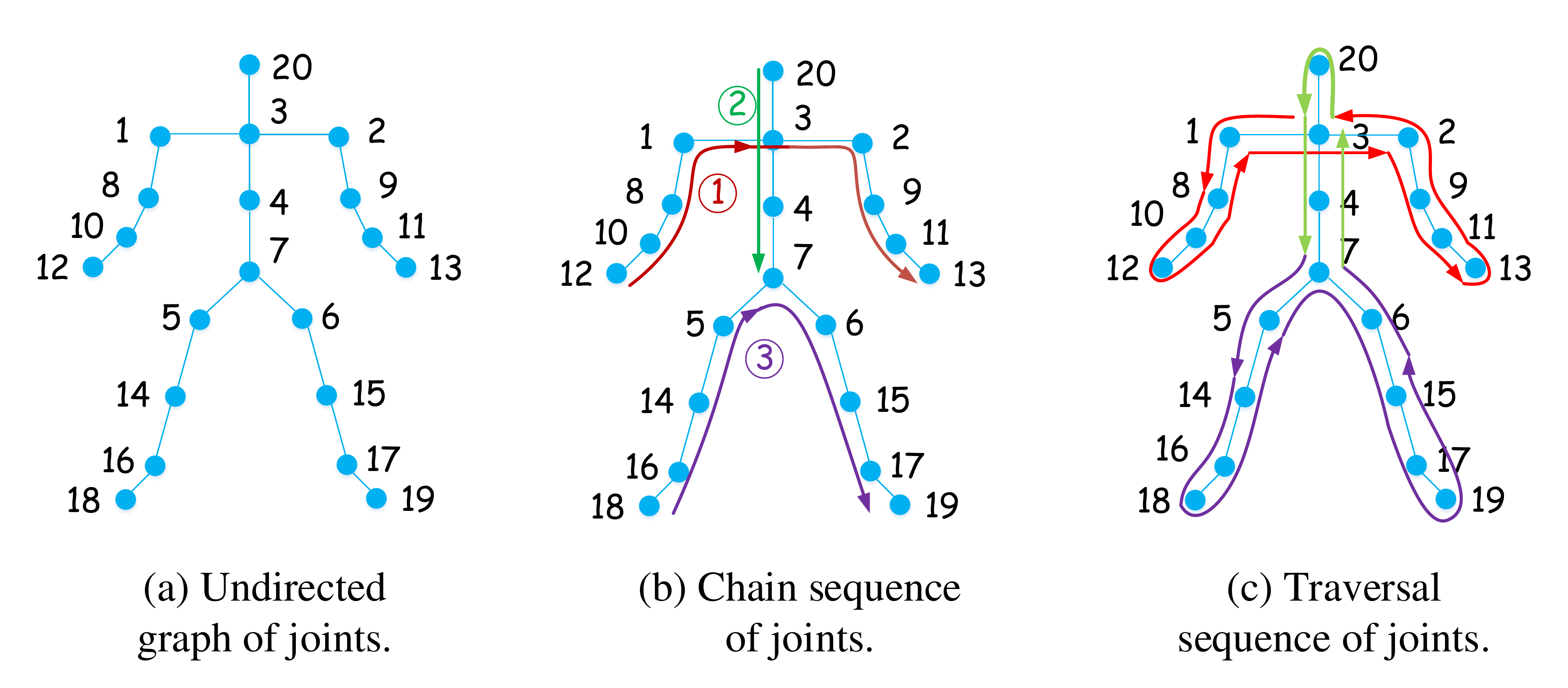}
\caption{(a) The physical structure of 20 joints. (b) Convert the joints graph into a sequence. The joints of arms come first, then that of body, finally is that of legs. (c) Use a traversal method to transform the joints graph into a sequence. The order of the sequence is the same as the visiting order of the arrow.}
\label{fig:skeleton_traveral}
\end{figure}

Human body can be considered as an articulated system of rigid segments connected by joints. Take the MSR Action3D dataset \cite{li2010action} as an example, the physical structure of the 20 joints is represented by an undirected graph in Figure \ref{fig:skeleton_traveral}(a). Nodes denote the joints and edges denote the physical connections. When an action takes place, this undirected graph displays some varied patterns of spatial structures. For example, clapping is performed with the joints of the two palms striking together, and bending is acted when the joints of the trunk shape into a curve.

To model the spatial dependency of joints, we cast the graph structure into a sequence of joints and exactly develop a relevant RNN architecture. The input of the RNN architecture at each step corresponds to the vector of coordinates of a certain joint.
As a joint has only three coordinates, we select a temporal window centered at the time step and concatenate the coordinates inside this window to represent this joint. This RNN architecture models the spatial relationships of joints in a graph structure and is called spatial RNN. The central problem is how to convert a graph into a sequence. We provide two alternative methods below.

\noindent \textbf{Chain sequence.} We assume the joints are arranged in a chain-like sequence with the order of arms, trunk and legs. The trunk is placed in the middle as it connects both arms and legs. For example, the 20 joints graph of the MSR Action3D dataset is arranged in a chain sequence in Figure \ref{fig:skeleton_traveral}(b). The \emph{chain sequence} maintains the physical connections of joints of each body part (arms, trunk and legs), and the joints are placed in a sequence without duplication. One of the drawbacks is that there is no physical connections at the boundary of joints between hands, trunk and legs. For instance, the joint whose index is 13 is not connected with the joint whose index is 20. But the two joints are adjacent in the generated chain-like sequence.

\noindent \textbf{Traversal sequence.} To address the limitation of the \emph{chain sequence}, we propose a graph traversal method to visit the joints in a sequence in the light of the adjacency relations, partly inspired by the tree-structure based traversal method \cite{liu2016spatio}.As illustrated in Figure \ref{fig:skeleton_traveral}(c), we first select the central spine joint as the starting point, and visit the joints of the left arm. While reaching an end point, it goes back. Then we visit the right arm, the upper trunk, etc. After visiting all joints, it finally returns to the starting point. We arrange the graph into a sequence of joints according to the visiting order. The \emph{traversal sequence} guarantees the spatial relationships in a graph by accessing most joints twice in both forward and reverse directions.

Different from the temporal RNN, spatial RNN could recognize actions by a glimpse of one frame (when the size of temporal window equals 1). Here, we do not use a hierarchical structure based on body parts, as the number of joints is limited (e.g., 25 for the NTU RGB+D dataset).

\subsection{3D transformation of skeletons}
For skeleton based action recognition, the input data is a sequence 3D coordinates of joints. As neural networks often require a lot of data to improve generalization and prevent overfitting, we exploit several data augmentation techniques based on 3D transformation to make the best use of limited supply of training data. Note that the 3D transformation techniques are only used during training.

\noindent \textbf{Rotation.} Based on Euler's rotation theorem, any 3D rotation can be given as a composition of rotations about three axes. The three basic rotation matrices in terms of rotate angles \(\alpha ,\beta ,\gamma \) about the \(x,y,z\) axis in a counterclockwise direction are represented as below:
\begin{equation}
{R_x}(\alpha ) = \left[ {\begin{array}{*{20}{c}}
1&0&0\\
0&{\cos \alpha }&{ - \sin \alpha }\\
0&{\sin \alpha }&{\cos \alpha }
\end{array}} \right]
\end{equation}
\begin{equation}
{R_y}(\beta ) = \left[ {\begin{array}{*{20}{c}}
{\cos \beta }&0&{\sin \beta }\\
0&1&0\\
{ - \sin \beta }&0&{\cos \beta }
\end{array}} \right]
\end{equation}
\begin{equation}
{R_z}(\gamma ) = \left[ {\begin{array}{*{20}{c}}
{\cos \gamma }&{ - \sin \gamma }&0\\
{\sin \gamma }&{\cos \gamma }&0\\
0&0&1
\end{array}} \right]
\end{equation}

General rotations can be obtained from these three basic rotation matrices using matrix multiplication:
\begin{equation}
R = {R_z}(\gamma ){R_y}(\beta ){R_x}(\alpha )
\end{equation}
where \(R\) is the general rotation matrix in the 3D coordinate system.

For the 3D coordinates of joints, we randomly rotate the input sequence of skeletons within a certain range for the \(x,y\) axis, as the rotation plane of the camera is perpendicular to the \(z\) axis. The rotation transformation simulates the viewpoint changes of the camera and improves the robustness of our model for cross view experimental settings. We find the recent work \cite{du2016representation} also uses the rotation transformation for cross view recognition of actions.

\noindent \textbf{Scaling.} Scaling transformation is used to change the size of skeletons. The transformation matrix can be formulated as:
\begin{equation}
S = \left[ {\begin{array}{*{20}{c}}
{{s_x}}&0&0\\
0&{{s_y}}&0\\
0&0&{{s_z}}
\end{array}} \right]
\end{equation}
where \({s_x},{s_y},{s_z}\) are scaling factors along with the three axes, respectively.

The scaling transformation can either expand or compress the dimensions of skeletons by using random scaling factors. As different action performers have varied heights and body sizes, the dimensions of their skeletons may be different. Thus the scaling transformation is beneficial for cross subject experimental settings.

\noindent \textbf{Shear.} Shear transformation is a linear map that displaces each point in a fixed direction. It slants the shape of the coordinates of joints and changes the angles between them. The transformation matrix can be represented as below:
\begin{equation}
Sh = \left[ {\begin{array}{*{20}{c}}
1&{sh_x^y}&{sh_x^z}\\
{sh_y^x}&1&{sh_y^z}\\
{sh_z^x}&{sh_z^y}&1
\end{array}} \right]
\end{equation}
where \({sh_x^y},{sh_x^z},{sh_y^x},{sh_y^z},{sh_z^x},{sh_z^y}\) are shear factors.

\section{Experiments}
The proposed model is evaluated on three datasets: NTU RGB+D dataset \cite{shahroudy2016ntu}, SBU Interaction dataset \cite{yun2012two}, and ChaLearn Gesture Recognition dataset \cite{escalera2013multi-modal,escalera2014chalearn}.

\subsection{Datasets}
\noindent \textbf{NTU RGB+D dataset}. Currently, this is the largest depth-based action recognition dataset, providing 3D coordinates of 25 joints collected by Kincet v2. It contains more than 56 thousand sequences and 4 million frames, captured in various background conditions. The dataset has 60 different action classes including daily, mutual, and health-related actions. The actions are performed by 40 different human subjects, whose age range is from 10 to 35.
Numerous variations in subjects and views, and large amount of samples make it highly suitable for deep learning methods. We follow the cross subject and cross view evaluations \cite{shahroudy2016ntu} and report the classification accuracy in percentage.

\noindent \textbf{SBU Interaction dataset}. This is a complex human activity dataset depicting two person interactions captured with Kinect. Each skeleton has 15 joints. It includes 282 skeleton sequences in 6822 frames. All videos are recorded in the same laboratory environment with 8 activities performed by 7 participants. The dataset is very challenging because the interactions are non-periodic, and have very similar body movements. Following the 5-fold cross validation \cite{yun2012two}, we split the 21 sets of this dataset into 5 folds and give the average recognition accuracy.


\noindent \textbf{ChaLearn Gesture Recognition dataset}. This dataset contains 20 Italian gestures performed by 27 different persons. There are 23 hours of Kinect data, consisting of RGB, depth, foreground segmentation and skeletons. The dataset has 955 videos in total. Each video lasts 1 to 2 minutes and contains 8 to 20 noncontinuous gestures. Here, we only use skeletons for gesture recognition. As done in the literature \cite{escalera2013multi-modal,fernando2015modeling}, we report the precision, recall and F1-score measures on the validation set.

\subsection{Implementation details} \label{implement_details}
We normalize skeletons by subtracting the central joint, which is the average of 3D coordinates of the hip center, hip left and hip right. The sequences are converted to a fixed length \(T\) by sampling and zero padding to allow for batch learning. \(T\) should be larger than the length of most sequences to reduce loss of information caused by sampling.

The NTU RGB+D dataset has a variable (one or two) number of persons performing actions. For samples with two persons, we only process one sequence each time, and average the predicted scores of the two. We set \(T = 100\) for this dataset, as most sequences are less than 100 in length. For the SBU Interaction dataset with a pair of skeletons representing interactions of two persons, we concatenate the two 3D coordinates for each joint at each time step and regard it as one sequence of 6D coordinates. We set the normalized sequence length \(T = 35\) for this dataset. For the ChaLearn Gesture Recognition dataset, we set \(T = 50\).

For the NTU RGB+D dataset, the number of neurons of each layer of \emph{stacked RNN} is 512. For \emph{hierarchical RNN}, the number of neurons of the body part and the whole body are 128 and 512, respectively. For the ChaLearn Gesture Recognition dataset, the networks structures are the same as those of the NTU RGB+D dataset. Compared with the above two datasets, the SBU Interaction dataset has less number of training samples and the sequence length is shorter. So we reduce the number of neurons of \emph{stacked RNN} of the temporal RNN to 256, and set the number of neurons of the body part and the whole body to 64 and 256, respectively. For all the datasets, the structure of the spatial RNN is the same as that of \emph{stacked RNN} of the temporal RNN. We adopt LSTM neurons for all layers due to its excellent performance for sequence recognition.

To demonstrate the effectiveness of the two-stream RNN, we simply adopt \emph{stacked RNN} for the temporal channel and \emph{chain sequence} for the spatial channel.
The weight of predicted scores of the temporal RNN is 0.9, and the temporal window size of the spatial RNN is one fourth of the fixed length \(T\), both are determined by cross-validation. The networks are trained using stochastic gradient descent. The learning rate, initiated with 0.02, is reduced by multiplying it by 0.7 every 60 epochs during training. The implementation is based on Theano \cite{2016arXiv160502688short} and Lasagne \footnote{\url{https://github.com/Lasagne/Lasagne}}. One NVIDIA TITAN X GPU is used to run all experiments.

\begin{table*}[!tbp]
\vspace{-2mm}
  \centering
    \caption{Comprehensive evaluation results of two-stream RNN on three datasets.}
\vspace{1mm}
  \resizebox{!}{1.40cm}{
\begin{tabular}{c|c|cc|c|ccc}
 \thickhline
\multirow{2}[0]{*}{Channel} & \multirow{2}[0]{*}{(\%)} & \multicolumn{2}{c|}{NTU RGB+D} & \multirow{2}[0]{*}{SBU Interaction} & \multicolumn{3}{c}{ChaLearn Gesture} \\
\cline{3-4}
\cline{6-8}
      &       & Cross subject & Cross view &       & Precision & Recall & F1-score \\
\hline
\multirow{2}[0]{*}{Temporal RNN} & Stacked & 66.1  & 68.9  & 89.0  &  89.5     &    89.6   &  89.5 \\
      & Hierarchical & 67.8  & 70.5  & 90.2  &   89.8    &   89.9    &  89.7 \\
\hline
\multirow{2}[0]{*}{Spatial RNN} & Chain & 53.7  & 58.9  & 82.2  & 81.9  & 82.1  & 81.9 \\
      & Traversal & 55.2  & 60.5  & 86.6  & 84.0    & 84.2  & 84.0 \\
\hline
\multirow{2}[0]{*}{Two-stream RNN} & No transform & 68.6  & 71.7  & 91.9  &  91.3     &   91.3    &  91.3 \\
      & 3D transform & 71.3  & 79.5  & 94.8  &   91.7    &  91.8     &  91.7 \\
 \thickhline
\end{tabular}}
  \label{tab:exp_two_stream}
\end{table*}

\begin{figure*}
\centering
\vspace{-4mm}
\includegraphics[width=0.9\linewidth]{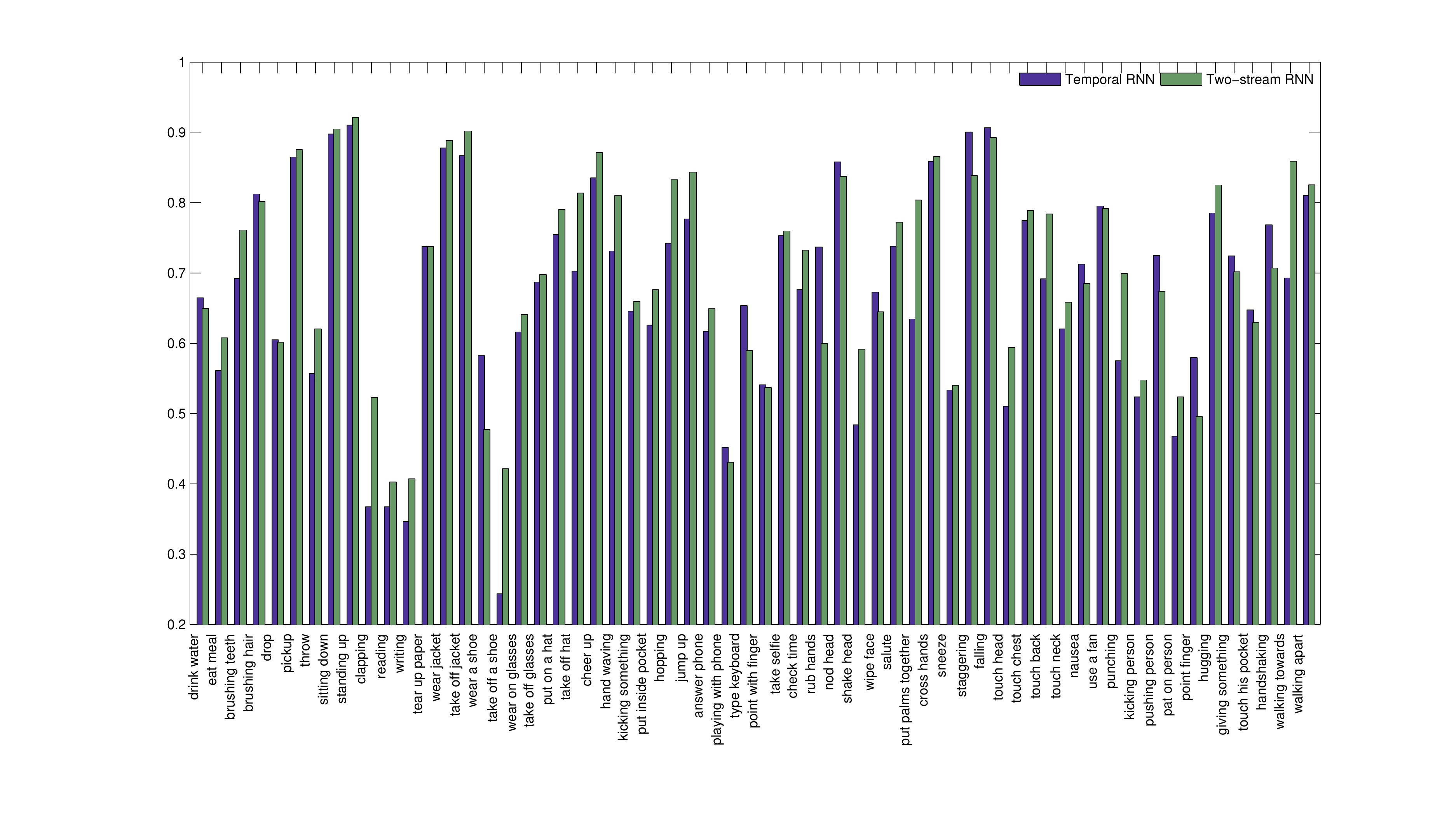}
\vspace{-9mm}
\caption{Accuracy for each action on the NTU RGB+D dataset.}
\label{fig:acc_per_ntu}
\vspace{-2mm}
\end{figure*}

\subsection{Experimental results}

\noindent \textbf{Comparison between models}. The comprehensive results of our two-stream RNN on three datasets are shown in Table \ref{tab:exp_two_stream}. We can see that the two-stream RNN consistently outperforms the individual temporal RNN and spatial RNN, which confirms that the spatial and temporal channels are both effective and complementary. In addition, for two activity recognition datasets, the 3D transformation techniques bring significant performance improvement for skeleton based recognition, especially for cross view evaluation. For example, on the NTU RGB+D dataset, the two-stream RNN with 3D transformation outperforms that without 3D transformation by \(7.8\% \) for cross view evaluation, much higher than the outperformed value of \(2.7\% \) for cross subject evaluation. The explanation is straightforward that rotation transformation randomly generates new skeletons from different views, thus making our two-stream RNN robust to the viewpoint changes.

Generally, the results of the temporal RNN are much better than those of the spatial RNN. This observation is consistent with the fact that most previous RNN based methods adopt the temporal RNN to recognize actions.
For the temporal RNN, the hierarchical structure generally performs better than the stacked structure. For example, on the NTU RGB+D dataset, \emph{hierarchical RNN} outperforms \emph{stacked RNN} by an average of 1.6\%.
For the spatial RNN, the results of the \emph{traversal sequence} are better than those of the \emph{chain sequence} as the traversal method maintains better spatial relationships of the graph structure by visiting most joints twice in both forward and reverse directions.

\noindent \textbf{Comparison between structures}. In Section \ref{implement_details} we manually define the structures of both \emph{stacked RNN} and \emph{hierarchical RNN}. Here we empirically study the effects of the number of stacked layers and the number of neurons for each layer on the performance. Due to the limited space, we only give results on the NTU RGB+D dataset by cross view protocol in Table \ref{tab:config_network}.

For \emph{stacked RNN}, we observe that two stacked layers (\emph{R512-512}) performs better than one layer (\emph{R512}), and three stacked layers (\emph{R512-512-512}) performs even better than two stacked layers. For the number of neurons of RNN layers, decreasing it to 256 (\emph{R256-256}) reduces the accuracy and increasing it to 1024 (\emph{R1024-1024}) does not necessarily improve the result. As adding more layers and increasing hidden neurons result more parameters and increase the computational complexity of our model, we adopt \emph{R512-512} as the default structure for \emph{stacked RNN}.

For \emph{hierarchical RNN}, using two stacked RNN layers for the part (\emph{P128-128, B512}) and increasing the number of neurons of the part from 128 to 256 (\emph{P256, B512}) improve the performances. The accuracy can be further improved by increasing the number of neurons of both the part and the whole body (\emph{P256, B1024}). To make a fair comparison with the stacked structure (\emph{R512-512}) and reduce the computational cost, we keep the structure with two layers and choose 128 as the number of neurons for the part, which is one fourth of the number of neurons for the whole body.


\begin{table}[!tbp]
  \centering
    \caption{Empirical study of networks structures. For \emph{stacked RNN}, \emph{R512-512} denotes two stacked layers of RNN with 512 hidden neurons. Similarly, \emph{R1024} denotes one RNN layer with 1024 hidden neurons. For \emph{hierarchical RNN}, \emph{P128-128, B512} denotes two stacked RNN layers with 128 hidden neurons for the body part and one RNN layer with 512 hidden neurons for the whole body. And so on for the other symbols. The default structures of \emph{stacked RNN} and \emph{hierarchical RNN} are \emph{R512-512} and \emph{P128, B512}, respectively.}
\vspace{1mm}
  \resizebox{!}{1.20cm}{
\begin{tabular}{c|c|c|c}
 \thickhline
\multicolumn{2}{c|}{Stacked RNN} & \multicolumn{2}{c}{Hierarchical RNN} \\
\hline
R512-512 & 68.9  & P128, B512 & 70.5 \\
\hline
R512-512-512 & 69.2  & P128-128, B512 & 71.4 \\
\hline
R512  & 68.6  & P256, B512 & 71.4 \\
\hline
R1024-1024 & 68.9  & P128, B1024 & 70.6 \\
\hline
R256-256 & 68.2  & P256, B1024 & 72.2 \\
 \thickhline
\end{tabular}}
  \label{tab:config_network}
\vspace{-2mm}
\end{table}

\subsection{Two-stream RNN versus temporal RNN}
As previous RNN based methods merely use the temporal RNN, here we aim to show the superiority of our two-stream RNN over the temporal RNN.

\begin{figure}
\centering
\includegraphics[width=0.9\linewidth]{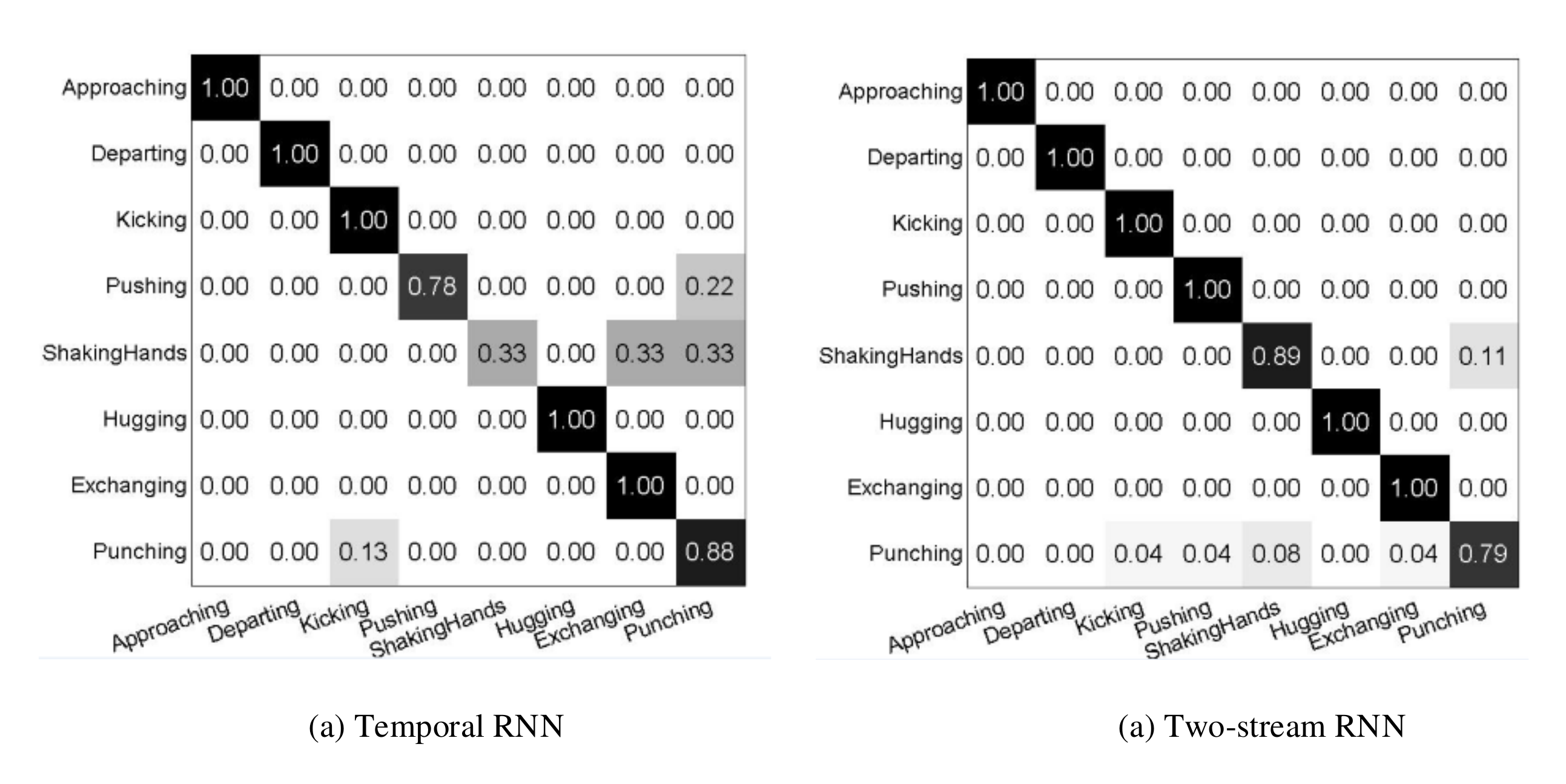}
\caption{Comparison of confusion matrices on the SBU Interaction dataset.}
\label{fig:confusion_mat}
\vspace{-2mm}
\end{figure}

We plot and compare the confusion matrices of our two-stream RNN and the temporal RNN on the SBU Interaction dataset in Figure \ref{fig:confusion_mat}. We can observe that there are three pairs of misclassified actions for the temporal RNN, but only one pair for our two-stream RNN. Moreover, for pushing, the samples are \(22\% \) misclassified as punching by the temporal RNN, while our two-stream RNN can correctly recognize all the samples.

We also depict the accuracy of each action. Figure \ref{fig:acc_per_ntu} shows the results of cross subject evaluation on the NTU RGB+D dataset.
For most actions, the accuracy of our two-stream RNN is higher than that of the temporal RNN.
For example, for brushing teeth, shaking head, and walking towards, the accuracy of the two-stream RNN is more than \(8\% \) higher than that of the temporal RNN.


\subsection{Parameter sensitivity}
\begin{figure}
  \centering
  \subfigure[Weight of temporal RNN]{
    \label{fig:param_analysis:a}
    \includegraphics[width=1.45in]{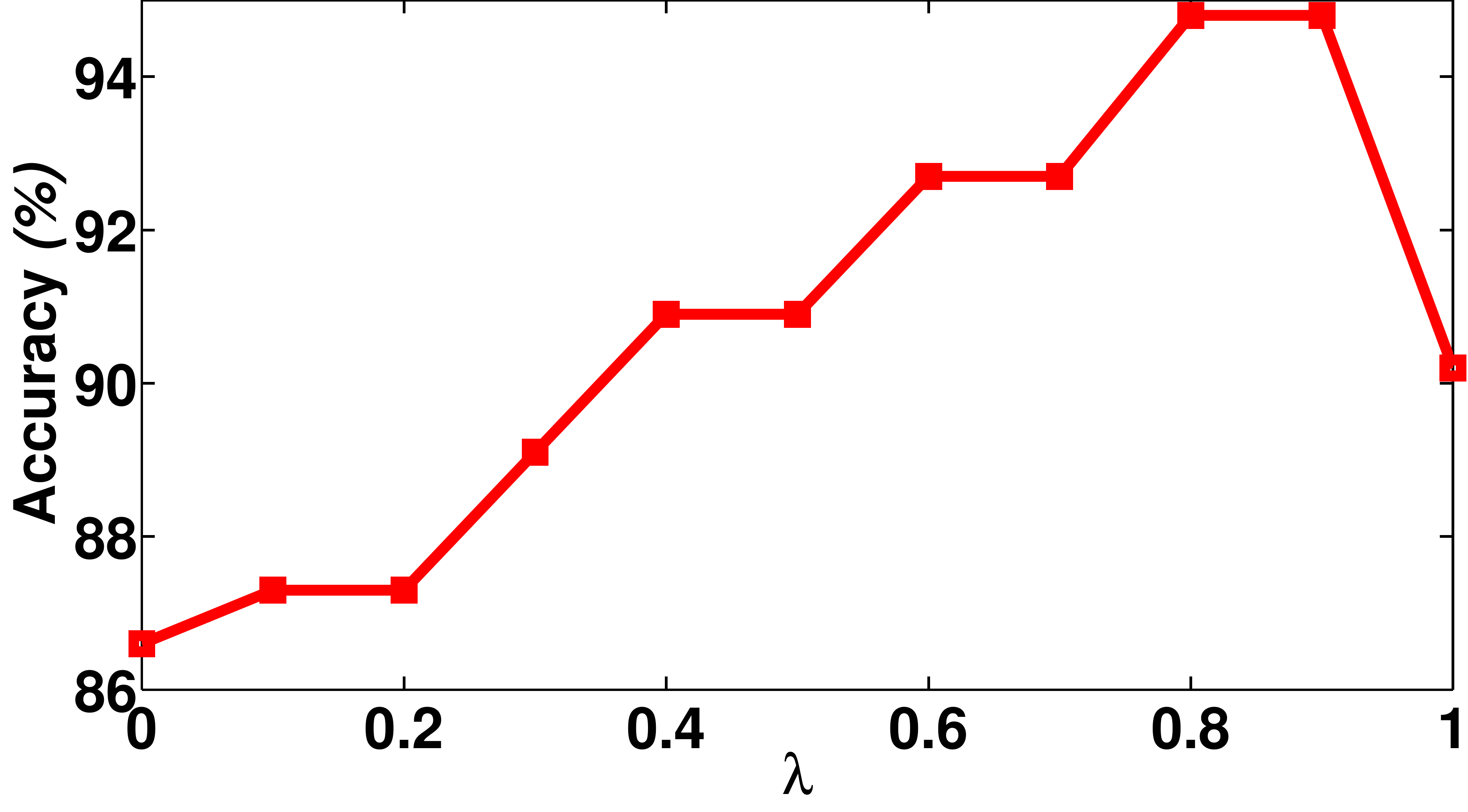}}
  \hspace{0.00in}
  \subfigure[Temporal window size]{
    \label{fig:param_analysis:b}
    \includegraphics[width=1.45in]{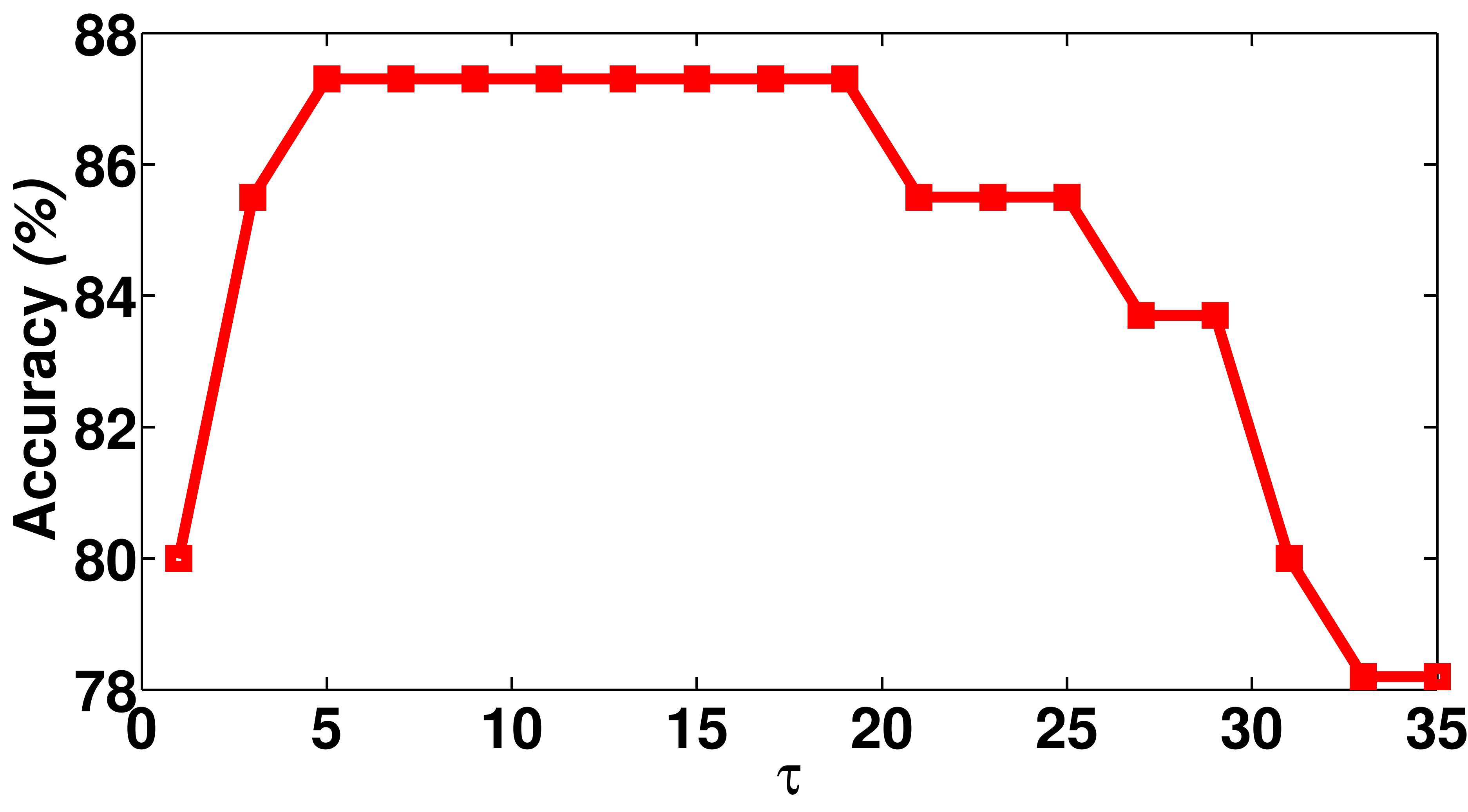}}
  \caption{Parameter sensitivity analysis on the SBU Interaction dataset. Here \(0 \le \lambda  \le 1\) and \(1 \le \tau  \le T\), where \(T=35\) is the sequence length after preprocessing.}
  \label{fig:param_analysis}
\vspace{-2mm}
\end{figure}

In this section, we evaluate the impact of parameters on the performance. Our two-stream RNN has two parameters, i.e., the size of temporal window of the spatial channel, and the weight of the temporal channel, denoted by \(\lambda \) and \(\tau \), respectively.
Figure \ref{fig:param_analysis} shows the evaluation results on the SBU Interaction dataset. It should be noted that similar results are observed for other datasets.

Figure \ref{fig:param_analysis}(a) shows the accuracy of the two-stream RNN w.r.t. the parameter \(\lambda \), \(\lambda  \in {\rm{\{ 0}},{\rm{0}}{\rm{.1}}, \cdots ,{\rm{0}}{\rm{.9}},{\rm{1\} }}\). We can see the best performance is reached when \(\lambda {\rm{ = }}0.8\) or \(\lambda {\rm{ = }}0.9\). When \(\lambda  < 0.8\), the accuracy decreases with a smaller value of \(\lambda \). The best result is much higher than the two extreme points where \(\lambda  \in \{ 0,1\} \), which correspond to the spatial and temporal RNN, respectively.

We choose \(\tau  \in \{ 1,3,5, \cdots ,T\} \) and plot the accuracy of the spatial RNN in Figure \ref{fig:param_analysis}(b).
We find that when \(5 \le \tau  \le 17\), i.e., \({T \mathord{\left/
 {\vphantom {T 7}} \right.
 \kern-\nulldelimiterspace} 7} \le \tau  \le {T \mathord{\left/
 {\vphantom {T 2}} \right.
 \kern-\nulldelimiterspace} 2}\), the temporal RNN obtains the best result. The performance drops when \(\tau \) is not in this range.
We conclude that our result is not sensitive to \(\tau \) for a wide range.

\subsection{Comparison with the state-of-the-art}
We compare our two-stream RNN method with the recent methods in the literature. Table \ref{tab:state_of_art_ntu} shows the results on the NTU RGB+D dataset.
We first compare our method with three traditional methods, i.e., 3D skeletons representation in a Lie group \cite{vemulapalli2014human}, Fisher vector encoding of \emph{skeletal quads} \cite{evangelidis2014skeletal} and \emph{FTP dynamic} \cite{hu2015jointly}. We observe that our performances are significantly higher, which shows the superiority of deep learning methods over the methods based on handcrafted features.
Then our method is compared with other deep learning methods based on RNN. Our results are much better than the reported results of \emph{HBRNN} \cite{du2015hierarchical} and \emph{Part-aware LSTM} \cite{shahroudy2016ntu}, both of which only model temporal dynamics of actions. Moreover, our method outperforms the newest spatio-temporal LSTM with trust gates \cite{liu2016spatio} by \(2.1\% \) and \(1.8\% \) for both cross subject evaluation and cross view evaluation, respectively.

The results on the SBU Interaction dataset are shown in Table \ref{tab:state_of_art_sbu}. Our result is \(7.9\% \) higher than the best result based on handcrafted features (Joint Feature \cite{ji2014interactive}). In addition, our approach is superior than recent RNN based approaches by outperforming the existing best result by \(1.5\% \). This experiment demonstrates our two-stream RNN model can recognize interactions performed by two persons very well.

The results on the Chalearn Gesture Recognition dataset are summarized in Table \ref{tab:state_of_art_chalearn}. Here our two-stream RNN is only compared with the methods solely based on skeletons. For precision, recall and F1-score, our approach yields state-of-the-art performance, outperforming the recently proposed VideoDarwin \cite{fernando2015modeling} by more than \(16\% \).

\begin{table}[!tbp]
  \centering
    \caption{Comparison of the proposed approach with the state-of-the-art methods on the NTU RGB+D dataset.}
\vspace{1mm}
  \resizebox{!}{1.60cm}{
\begin{tabular}{c|c|c}
 \thickhline
Method & Cross subject & Cross view \\
\hline
Lie Group \cite{vemulapalli2014human} & 50.1  & 52.8 \\
Skeletal Quads \cite{evangelidis2014skeletal} & 38.6  & 41.4 \\
FTP Dynamic \cite{hu2015jointly} & 60.2  & 65.2 \\
HBRNN \cite{du2015hierarchical} & 59.1  & 64.0 \\
Part-aware LSTM \cite{shahroudy2016ntu} & 62.9  & 70.3 \\
Trust Gate ST-LSTM \cite{liu2016spatio} & 69.2  & 77.7 \\
\hline
Two-stream RNN & \textbf{71.3}  & \textbf{79.5} \\
 \thickhline
\end{tabular}}
  \label{tab:state_of_art_ntu}
\vspace{-2mm}
\end{table}

\begin{table}[!tbp]
  \centering
    \caption{Comparison of the proposed approach with the state-of-the-art methods on the SBU Interaction dataset.}
\vspace{1mm}
  \resizebox{!}{1.60cm}{
\begin{tabular}{c|c}
 \thickhline
Method & Accuracy \\
\hline
Joint Feature \cite{yun2012two} & 80.3 \\
Joint Feature \cite{ji2014interactive} & 86.9 \\
HBRNN \cite{du2015hierarchical} & 80.4 \\
Deep LSTM \cite{zhu2016co} & 86.0 \\
Co-occurrence LSTM \cite{zhu2016co} & 90.4 \\
Trust Gate ST-LSTM \cite{liu2016spatio} & 93.3 \\
\hline
Two-stream RNN & \textbf{94.8} \\
 \thickhline
\end{tabular}}
  \label{tab:state_of_art_sbu}
\vspace{-2mm}
\end{table}

\begin{table}[!tbp]
  \centering
    \caption{Comparison of the proposed approach with the state-of-the-art methods on the ChaLearn Gesture Recognition dataset.}
\vspace{1mm}
  \resizebox{!}{1.55cm}{
\begin{tabular}{c|c|c|c}
 \thickhline
Method & Precision & Recall & F1-score \\
\hline
Skeleton Feature \cite{wu2013fusing} & 59.9  & 59.3  & 59.6 \\
Portfolios \cite{yao2014gesture} &  --     &   --    & 56.0 \\
Gesture Spotting \cite{pfister2014domain-adaptive} & 61.2  & 62.3  & 61.7 \\
HiVideoDarwin \cite{wang2015hierarchical} & 74.9 & 75.6 & 74.6 \\ 
CNN for Skeleton \cite{du2015skeleton} & 91.2 & 91.3 & 91.2 \\
VideoDarwin \cite{fernando2015modeling} & 75.3  & 75.1  & 75.2 \\
\hline
Two-stream RNN & \textbf{91.7} & \textbf{91.8} & \textbf{91.7} \\
 \thickhline
\end{tabular}}
 \label{tab:state_of_art_chalearn}
\vspace{-2mm}
\end{table}

\section{Conclusion}
In this paper, we have proposed an end-to-end two-stream RNN architecture for skeleton based action recognition, with the temporal stream modeling temporal dynamics and the spatial stream processing spatial configurations. We explore two structures to model the sequence of joints of skeletons for the temporal stream. For the spatial stream, we also devise two methods to convert the structure of skeleton into a sequence before using a RNN to handle the spatial dependency. Moreover, to improve generalization and prevent overfitting for deep learning based methods, we employ rotation transformation, scaling transformation and shear transformation as data augmentation techniques based on 3D transformation of skeletons. Our experiments have shown that two-stream RNN outperforms existing state-of-the-art skeleton based approaches on datasets for generic actions (NTU RGB+D), interaction activities (SBU Interaction) and gestures (ChaLearn). In the future, we will consider to learn the structure patterns for the spatial channel and further improve the results.

\section*{Acknowledgement}
This work is jointly supported by National Key Research and Development Program of China (2016YFB1001000), National Natural Science Foundation of China (61525306, 61633021, 61420106015) and Beijing Natural Science Foundation (4162058). This work is also supported by grants from NVIDIA and the NVIDIA DGX-1 AI Supercomputer.

{\small
\bibliographystyle{ieee}
\bibliography{egbib}
}

\end{document}